\documentclass{article}

\usepackage{microtype}
\usepackage{graphicx}
\usepackage{booktabs} 

\usepackage{amsmath}
\usepackage{amssymb}
\usepackage{bm} 
\usepackage{subfig}
\usepackage{caption}
\usepackage{multirow}
\DeclareMathOperator*{\argmax}{argmax}

\usepackage{hyperref}

\usepackage[accepted]{icml2021}

\icmltitlerunning{Learning to Cooperate with Unseen Agents via Meta-Reinforcement Learning}

\begin{document}

\twocolumn[
\icmltitle{Learning to Cooperate with Unseen Agent via Meta-Reinforcement Learning}




\begin{icmlauthorlist}
\icmlauthor{Rujikorn Charakorn}{vistec}
\icmlauthor{Poramate Manoonpong}{vistec}
\icmlauthor{Nat Dilokthanakul}{vistec}
\end{icmlauthorlist}

\icmlaffiliation{vistec}{
Information Science and Technology, Vidyasirimedhi Institute of Science and Technology (VISTEC), Rayong, Thailand}

\icmlcorrespondingauthor{Nat Dilokthanakul}{natd\_pro@vistec.ac.th}

\icmlkeywords{Coordination, Meta-Learning, Reinforcement Learning, Ad-hoc Teamwork, Generalization, Multi-Agent System}

\vskip 0.3in
]



\printAffiliationsAndNotice{} 

\begin{abstract}
Ad hoc teamwork problem describes situations where an agent has to cooperate with previously unseen agents to achieve a common goal. For an agent to be successful in these scenarios, it has to have a suitable cooperative skill. One could implement cooperative skills into an agent by using domain knowledge
to design the agent's behavior. However, in complex domains, domain knowledge might not be available. Therefore, it is worthwhile to explore how to directly learn cooperative skills from data. In this work, we apply meta-reinforcement learning (meta-RL) formulation in the context of the ad hoc teamwork problem. Our empirical results show that such a method could produce robust cooperative agents in two cooperative environments with different cooperative circumstances: social compliance and language interpretation. (This is a full paper of the extended abstract version \citeauthor{charakorn2021learning}, \citeyear{charakorn2021learning}.)


\end{abstract}

\section{Introduction} \label{intro}

The problem of cooperation without prior coordination is known as the ad hoc teamwork problem \cite{ barrett2017making,mirsky2020penny,stone2010ad}. It has only been recently tackled in the field of multi-agent deep reinforcement learning (MADRL), where the problem seems to be exacerbated with the use of deep neural network models and their complex learning dynamics. Leading the way, \citet{bard2020hanabi} present a game environment that aims to accelerate the progress in this direction. They clearly highlight the failure of contemporary RL in the ad hoc teamwork problem. Then, \citet{hu2020other} propose Other-Play, a state-of-the-art method that exploits the known symmetry of the environment, which allows agents to cooperate robustly with unseen agents. However, these unseen agents need to also be trained with Other-Play, limiting the applicability of the method.

In this work, we posit that a cooperative agent needs the ability to generalize its policy to the dynamics of unseen agents. The generalization problem, in the single-agent domain, is being tackled from many different directions, e.g. using domain randomization \cite{andrychowicz2020dactyl,mehta2020adr,tobin2017dr}, model-based learning \cite{hafner2019dream,sekar2020planning}, and meta-learning \cite{finn2017model,mishra2017snail}. A notable direction is the use of the meta-learning paradigm. This paradigm involves learning an adaptive behavior using data. At test time, the agent can utilize newly collected samples to shape its behavior in an unseen environment. Intuitively, this method could work well in the ad hoc teamwork problem, where we would like our agent to adapt its behavior such that it cooperates smoothly with an unseen partner.

\citet{duan2016rl} and \citet{wang2016meta_rl} describe how meta-RL could be implemented using a recurrent neural network (RNN) in the single-agent domain. In their work, the RNN is used to represent \emph{a fast learning program} within its weights.
We are interested in using the same method to train an RNN to represent a \emph{cooperative program}. We argue that the main characteristic of a cooperative skill is the ability to adapt its policy by observing the current partner's behavior and then selecting suitable responses. Therefore, we hypothesize that a meta-RL algorithm is a suitable method to encode the cooperative skill into an agent \emph{without} using any domain knowledge to design the agent's behavior. This approach is useful when working with complex environments where domain knowledge might not be available. Instead of using domain knowledge to design cooperative behaviors, we move the engineering effort to partner generation and the training method instead. This is similar to how meta-learning moves engineering consideration from the design of learning algorithm to the design of task distribution.



Despite many studies of meta-RL, to the best of our knowledge, there has not been any documented report that investigates the applicability of meta-RL under the ad hoc teamwork problem. We believe this work is a worthwhile investigation that benefits both meta-RL and cooperative multi-agent learning communities. To further our understanding of this topic, we investigate the behavior of agents trained with meta-RL in the tasks that require cooperation. The key insight is that meta-RL can produce cooperative agents that can generalize to work with unseen agents. We show that a meta-RL agent can perform robustly under unexpected situations: longer horizon and partner switching without explicitly designed to work in those circumstances.
Furthermore, we examine two factors that could affect the adaptability of the meta-trained agent. We find that the quantity and diversity of training partners are both crucial for a meta-RL agent to generalize to unseen partners. Finally, we show that meta-RL agents can coordinate with each other as well as agents trained with different training algorithms. Importantly, they can cooperate with unseen learned partner agents despite only trained with pre-programmed partners.


\section{Related Work} \label{related_work}

\subsection{Ad Hoc Teamwork Problem}
Ad hoc teamwork problem has been extensively investigated \cite{mirsky2020penny, stone2010ad}. To solve this problem, most of the previous works are based on the idea of reasoning about other agents \cite{albrecht2018agent_modeling_review}. They are implemented either by explicit modeling of the partners' current policy \cite{barrett2011empirical,barrett2012learning} or using a classification model for selecting their roles \cite{barrett2017making,canaan2020generating}. These methods rely heavily on domain knowledge to design an agent that has desired cooperative skills.

In contrast, our approach uses meta-RL to train cooperative agents, which allow them to \emph{learn} about other agents by themselves at test time, without explicit programming of the reasoning process. This training paradigm is appealing because it requires less engineering effort during the training of the agent (i.e. it can be applied to different environments without requiring customization of the learning algorithm to the task at hand). Also, the agents do not need to know a priori the distribution of tasks, partner's role, or communication protocol.

\subsection{Meta-Reinforcement Learning}
Meta-Reinforcement Learning can be divided into the algorithms that require gradient descent update at the adaptation phase \cite{clavera2018learning,finn2017model,rajeswaran2019meta} and the algorithms that can be adaptive through information from new samples and internal dynamics of the model itself \cite{duan2016rl,mishra2017snail,wang2016meta_rl}. We will refer to these two types of meta-RL as gradient-based meta-RL and context-based meta-RL.  

The gradient-based meta-reinforcement learning, e.g. model-agnostic meta-learning (MAML) \cite{finn2017model}, trains the weights of a model to be easily adaptable via gradient descent. Therefore, in the adaptation phase, the model can adapt its behavior quickly through only a few gradient update steps. In contrast, context-based meta-RL adapts at every timestep by changing its internal dynamic without an explicit update procedure. We believe both approaches are worth investigating.
In this work, we start investigating the context-based method.

\subsection{Generalization Problem in MADRL}

In MADRL, agents are known for over-fitting to training partners making them unable to perform well when matched with unseen test-time agents \cite{carroll2019utility,lanctot2017unified}. Learned opponent model and policy representation \cite{he2016opponent,raileanu2018modeling,grover2018learning} has been studied in MADRL to enhance generalization when working with unseen agents. Also, those ideas from the single-agent domain have been applied to the multi-agent domain. 
Analogous to domain randomization, using a set of diverse training partners also improves an agent's ability to generalize \cite{bansal2017emergent,jaderberg2019human,vinyals2019grandmaster}. Additionally, if agents can be generated, one could use procedural generation to produce a set of diverse agents as training partners \cite{canaan2020generating,ghosh2019towards}.

Although adding diversity seems to be a trivial solution to the over-fitting problem, it is unclear how diversification should be applied in fully cooperative tasks. It has been shown that diversification methods that are expected to work in competitive games do not work in cooperative settings \cite{carroll2019utility,charakorn2020investigating}.

\section{Methods} \label{method}
\subsection{Formulation}
We formalize a fully cooperative environment as decentralized partially-observable Markov decision process (Dec-POMDP, \citealt{bernstein2002complexity}) which has a set of $N$ agents. At each timestep $t$ with state $s_t \sim \mathcal{S}$, agent $i$ observes $o^i_t \sim \mathcal{O}(o|s_t,i)$ and execute action $a^i_t$ sampled from stochastic policy $\pi^{i}(a^{i}_t|\tau^{i}_t)$ where $\tau^i_t$ contains the history of the current trajectory, observed by agent $i$ until timestep $t$, $\{o_0^i,a_0^i,...,o_t^i\}$. The state transition function is defined by $\mathcal{P}(s'|s_t,\bm{A})$ where $\bm{A}$ is the joint action of all agents. Since the game is fully cooperative, all agents receive the same reward $r_t \sim \mathcal{R}(r|s_t,\bm{A})$. The goal for all agents is to maximize their expected return:
\begin{gather} \label{expected_return}
\mathop{\mathbb{E}} [\sum_{t=0}^H \gamma^t \mathcal{R}(s_t,\bm{A} \sim (\pi^{1}(\tau^1_t),...,\pi^{n}(\tau^n_t)] \nonumber
\end{gather}
over time horizon $H$, discount factor $\gamma$.

\subsection{Context-Based Meta-Reinforcement Learning in the Multi-Agent Domain}
In general, meta-learning involves an update rule that enables neural networks to do fast learning when given a small dataset. Similarly for context-based meta-reinforcement learning, the agent is trained such that the neural network represents an adaptive policy that can adjust itself to a given context (i.e., an observed trajectory). By leveraging the training data from a distribution of tasks, the agent learns to exploit the structure of the training tasks. This information of task structure gives the agent useful inductive bias that allows it to adapt to new tasks quickly. Concretely, the meta-training objective for a meta-RL agent, in a single agent domain, is to find a policy $\pi^i$, parameterized by $\theta^\pi$, that maximizes the expected return in the training tasks distribution $\mathcal{D}$:
\begin{gather} \label{meta_obj_single}
    \argmax_{\theta^\pi} \mathop{\mathbb{E}_{d \sim \mathop{D}}}[\sum_{t=0}^H \gamma^t \mathcal{R}_d(s_t,a_t \sim \pi^i(\tau^i_t)] \nonumber 
\end{gather}
where $d$ is a training task sampled from $\mathcal{D}$ and $R_d$ is a reward function of task $d$. Essentially, this objective incentivizes the agent to learn a policy that performs well in every task using information from trajectory history $\tau^i_t$.

We are interested in applying context-based meta-reinforcement learning to the multi-agent domain for dealing with the ad hoc teamwork problem particularly. So, instead of using a distribution of tasks, we use a distribution of partner agents $\mathcal{P}$ in which each partner agent has a unique policy $\pi^{p}$. 
In this work, we only consider two-player cooperative games. Thus, we can write the meta-training objective in the multi-agent domain as follows:
\begin{gather}\label{meta_obj_multi}
    \argmax_{\theta^\pi} \mathop{\mathbb{E}_{p \sim \mathcal{P}}}[\sum_{t=0}^H \gamma^t \mathcal{R}(s_t,\bm{A} \sim (\pi^i(\tau^i_t), \pi^{p}(\tau^p_t))] \nonumber
\end{gather}
Still, meta-RL has some prerequisites in order to train an adaptive agent properly. One of them is specifying the right meta-training task distribution \cite{clune2019ai,finn2018metalearning,gupta2018unsupervised}. 
To achieve a diverse partner population, we use simple fixed behaviors to design partner agents. The benefits of this are threefold: (i) We can make sure that each agent in the population is unique and gives useful learning signals for the meta-learner. (ii) It sidesteps the non-stationary problem of learning in multi-agent RL. Consequently, we can use off-the-shelf RL algorithms to optimize the objective directly. (iii) We can control the diversity of the meta-training distribution and study the impact of diversity (in section \ref{diversity_test}).


Implementing context-based meta-RL can be done with recurrent neural networks or an attention mechanism, in combination with previous action and reward signal \cite{duan2016rl,mishra2017snail,wang2016meta_rl}. Using context-based meta-RL, the online adaptation process is learned and featured in the weights of the neural network.

\section{Cooperative Environments} \label{setup}
All experiments are carried out using two two-player cooperative environments with different cooperative circumstances. Every partner agent in these experiments has a pre-programmed behavior which we will describe in detail. These agents are used as training and test-time partners in every experiment\footnote{Partner agents in both environments could also be learned but it is non-trivial to obtain a diverse set of partners using off-the-shelf RL algorithms.} (except section \ref{adaptive_test} that uses learned test-time partners).

\subsection{Lever Coordination Game (LC Game)}
An important cooperative skill is the ability to adapt to different social norms and the ability to comply with social dynamics. To investigate this skill, we adapt the Lever coordination game from \citet{hu2020other}. 

The Lever coordination game consists of five levers; two players are playing at a time. At each timestep, each agent selects one of the levers. A positive reward of 1.0 is achieved if both of them select the same lever. However, an action cannot be repeated if it is already chosen until all the possible actions are taken. For example, at the start of a game, the action sequence [5,2,3,4,1] is valid while [3,4,2,1,3] is not (the last action must be replaced with 5).
After all the actions are taken, every action becomes available again.

Pre-programmed agents in this environment have a fixed pattern of actions, which keeps repeating a pattern of five actions. For example, one of the partner agents will keep choosing lever [2,4,3,5,1,2,4,3,5,1,...] and, then, keep repeating these actions. The observation for each agent in this game is only the current timestep of the trajectory since it is a stateless environment.


\subsection{Discrete Speaker-Listener Game (DSL Game)}
We adapt the speaker-listener game from \citet{lowe2017multi} to investigate the extent a meta-RL algorithm can create a good interpretation skill.\footnote{We found that the original version has a trivial solution for the listener which is to follow the reward gradient without listening to the speaker.} The game consists of a speaker, a listener, and five landmarks. At each timestep, a reward of 1.0 will be randomly assigned to one of the landmarks.
The speaker can see where the reward is but cannot get the reward by itself. The speaker will have to speak to the listener by emitting a one-hot vector to the listener. The listener then chooses one of the landmarks as a target. Both players get the reward if the listener selects the correct target.

Similar to the LC game, partner agents in this game are also pre-programmed. Depending on the assigned role, each partner agent will have a unique one-to-one mapping from messages to landmarks (as a listener) or from landmarks to messages (as a speaker). This means that each agent will use a different language depending on its mapping, essentially making them speak or interpret using different vocabulary.



\section{Experimental Results}
\begin{figure}[t]
    \centering
    \subfloat[RNN\label{rnn}]{
        \includegraphics[trim={12cm 3.9cm 12cm 3.9cm},clip,width=0.11\textwidth]{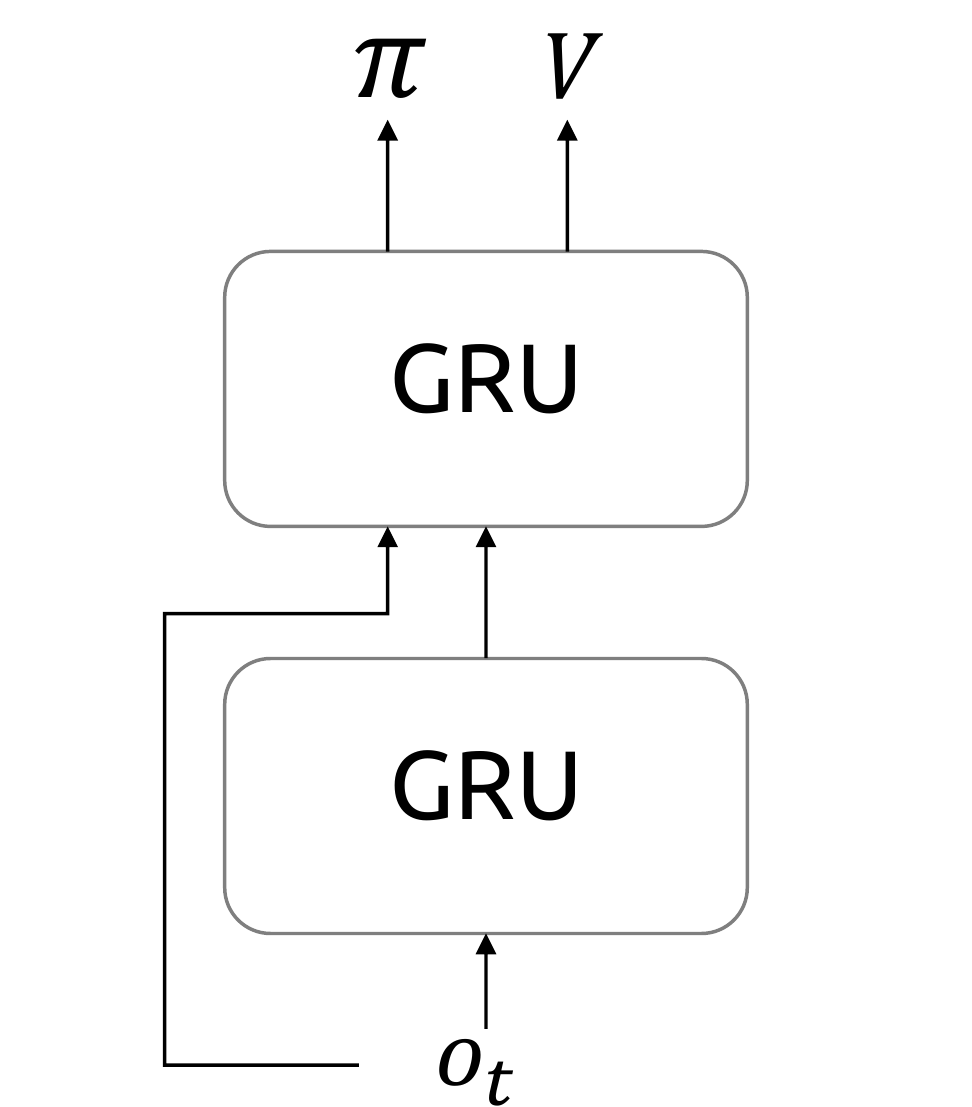}
    }
    \subfloat[a-RNN\label{a-rnn}]{
        \includegraphics[trim={12cm 3.9cm 12cm 3.9cm},clip,width=0.11\textwidth]{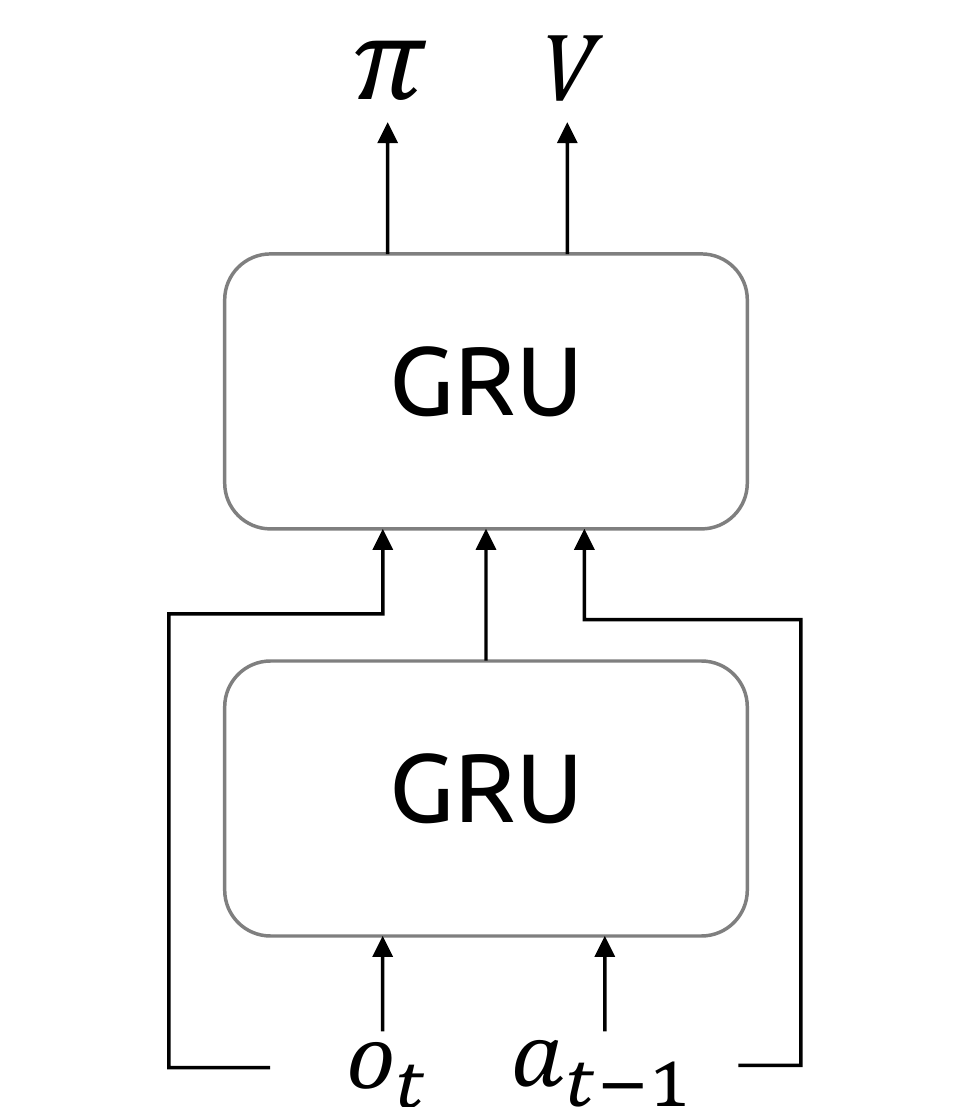}
    }
    \subfloat[r-RNN\label{r-rnn}]{
        \includegraphics[trim={12cm 3.9cm 12cm 3.9cm},clip,width=0.11\textwidth]{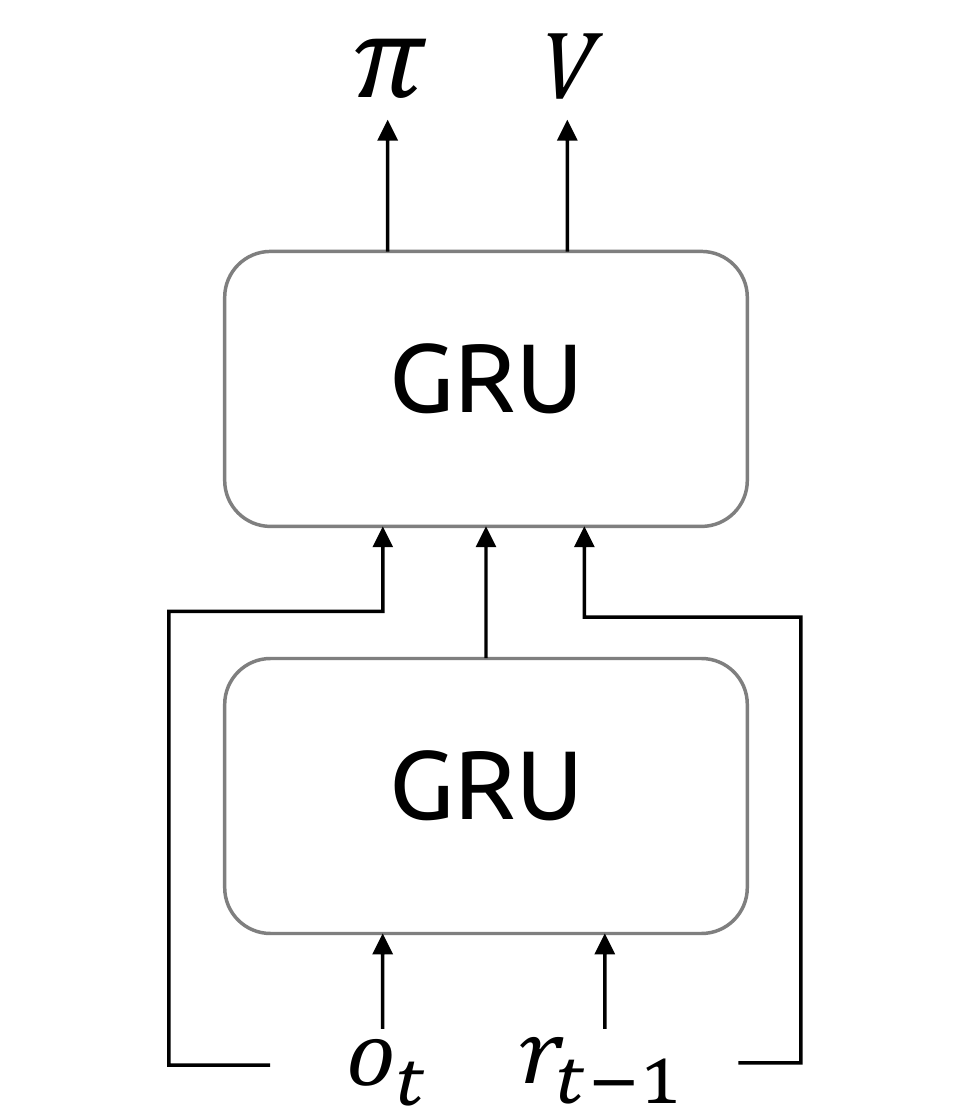}
    }
    \subfloat[ar-RNN\label{ar-rnn}]{
        \includegraphics[trim={12cm 3.9cm 12cm 3.9cm},clip,width=0.11\textwidth]{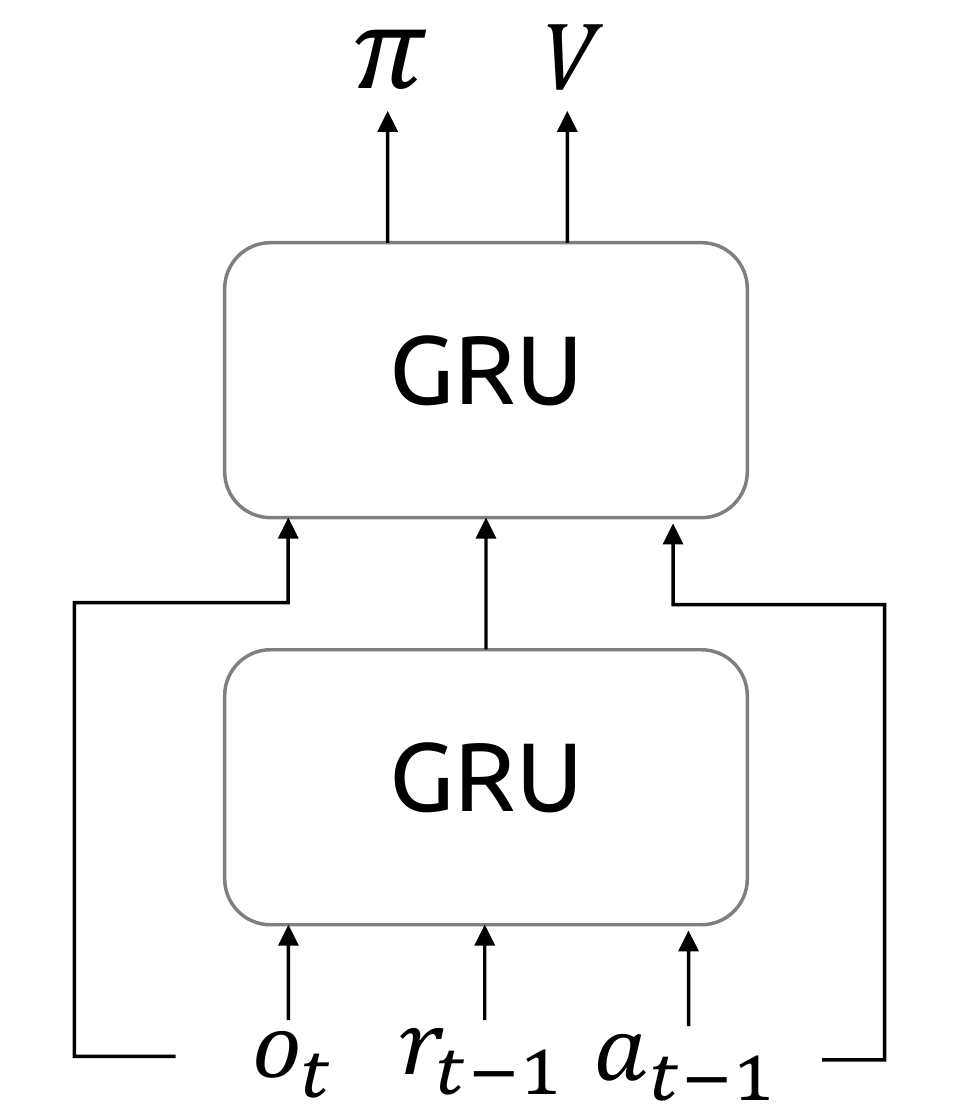}
    }

    \caption{\textbf{Architecture variations}. Four variations of architectures are considered in this work. Each architecture has different input features. These architectures are tested in the experiments to study the impact of each component of meta-RL.}
    \label{fig:architectures}
\end{figure}
    
Our experiments aim to answer the following questions: (i) Can meta-RL produce a cooperative agent? If so, which components of the meta-RL mechanism are necessary for the emergence of the cooperative skills? (ii) Can a meta-trained agent continually perform under a longer horizon and adapt to different partner agents?
(iii) How do quantity and diversity of training partners affect the generalization of a meta-RL agent?
(iv) can a meta-RL agent, which is trained with a non-adaptive agent, coordinate with unseen adaptive agents?

During the training, in a trajectory, the learning agent will be paired with a partner sample from the training distribution. It will interact with that partner until a new one is sampled at the beginning of a new trajectory. Thus, in each episode, the learning agent will interact with a different partner. There are 120 unique partners in both environments. We use 60 randomly chosen partners as training partners and 60 unseen agents as test-time partners. Every variation uses the same amount of timesteps during the training period (9M and 30M timesteps for the LC and DSL game respectively). 

Our evaluation protocol only allows adaptation within a trajectory. This is done by resetting the RNN states at the beginning of the trajectory. Similar to the training period, the agent will be matched with a new unseen agent sampled from the test-set at the beginning of each trajectory. An evaluation and training episode has an identical horizon length of 50 timesteps unless stated otherwise. 

\subsection{Emergence of Cooperative skills}
\begin{figure}[t]
    \centering
    \subfloat[LC Game]{
        \includegraphics[trim={0cm 1.5cm 0cm 1.5cm},clip,width=0.45\textwidth]{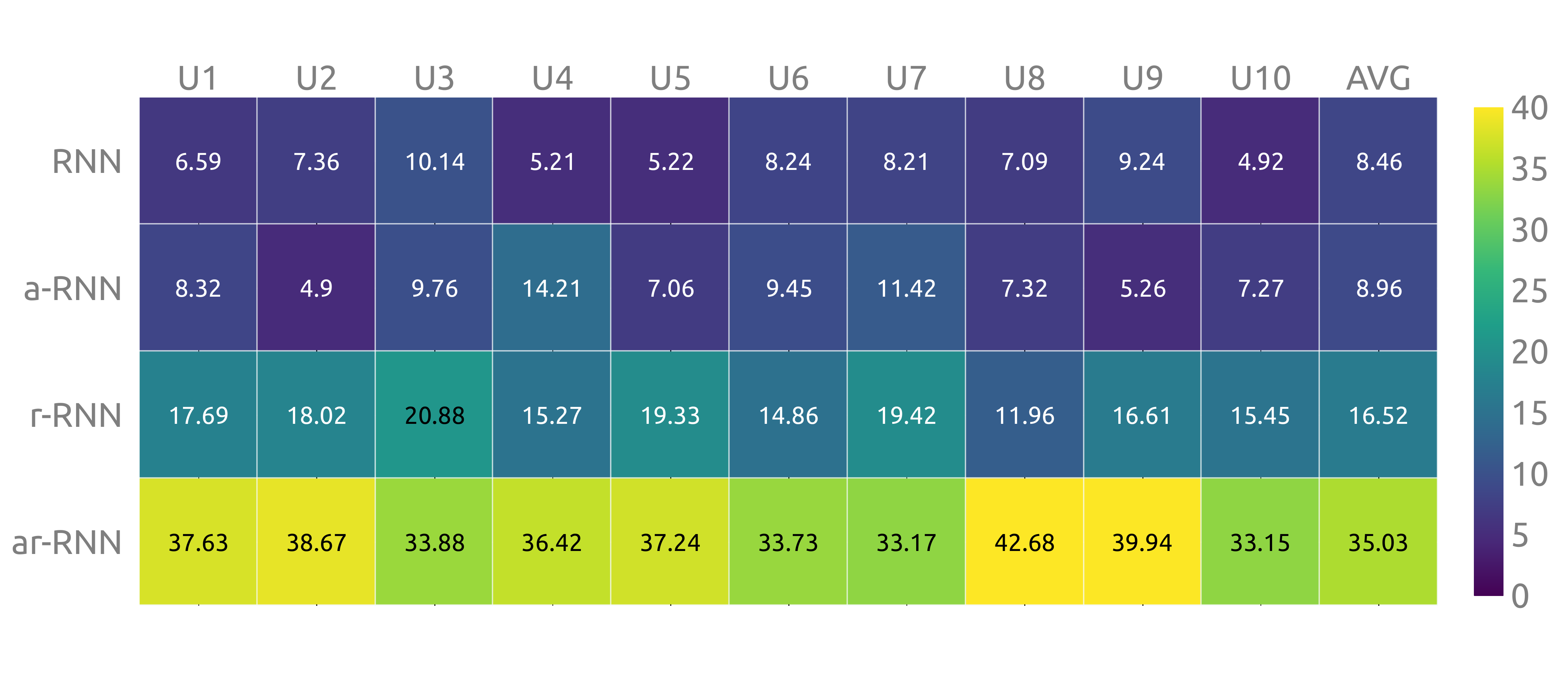}
    }
    \\
    \subfloat[DSL Game]{
        \includegraphics[trim={0cm 1.5cm 0cm 1.5cm},clip,width=0.45\textwidth]{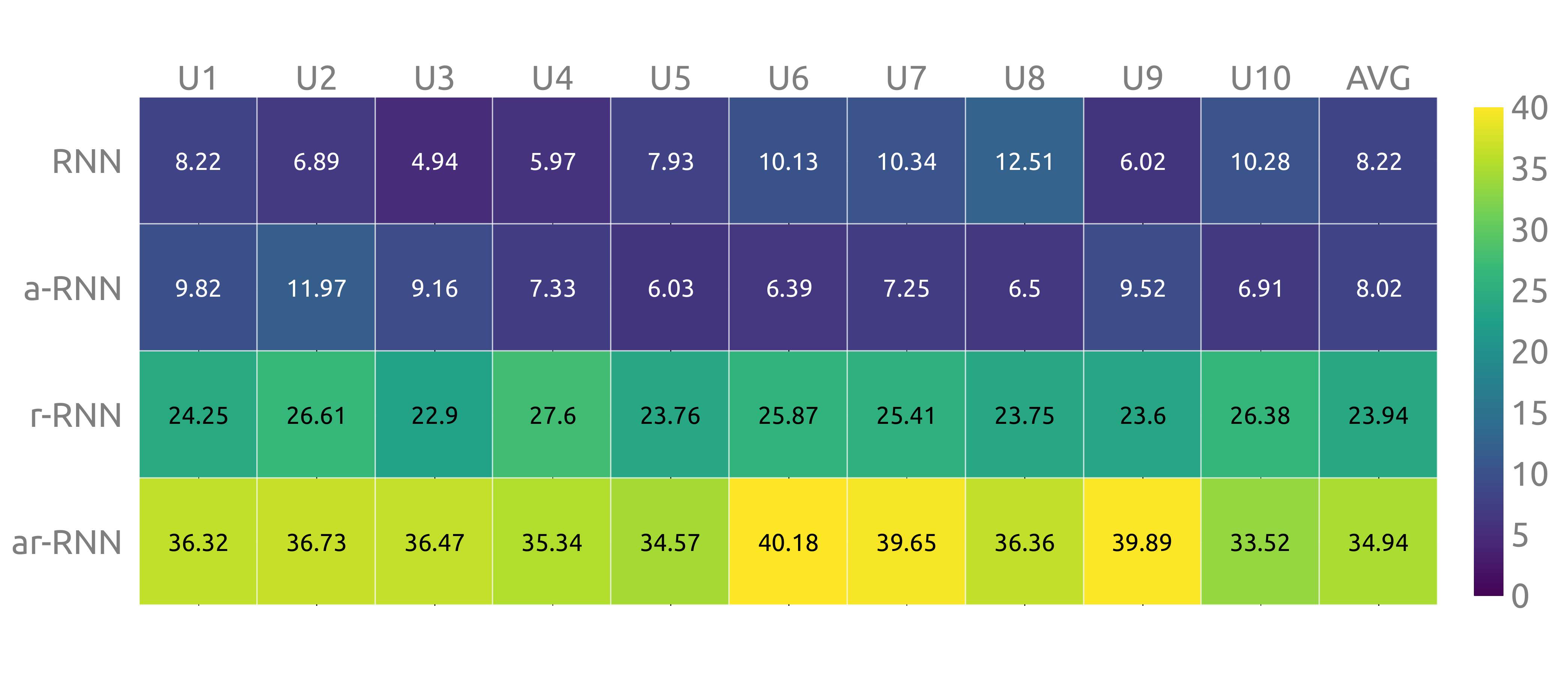}
    }

    \caption{\textbf{Test-time score when matched with unseen agents.} Each cell in the heatmap shows an average score over 5 training seeds for a pair of a trained agent and an unseen agent. The score is calculated by averaging the returns over evaluation episodes. For each training seed, we use a total of 18,000 evaluation episodes where the partner in each episode is randomly sampled from a pool of 60 test agents. Only 10 test agents are shown in the table due to the space constraint. The columns represent ID of test agents and the rows represent the neural network architecture. The average in the last column is calculated using all test agents.}
    \label{fig:unseen_table}
\end{figure}

\begin{figure}[t]
    \centering
    \subfloat[LC Game]{
        \includegraphics[trim={0cm 0cm 0cm 1cm},clip,width=0.45\textwidth]{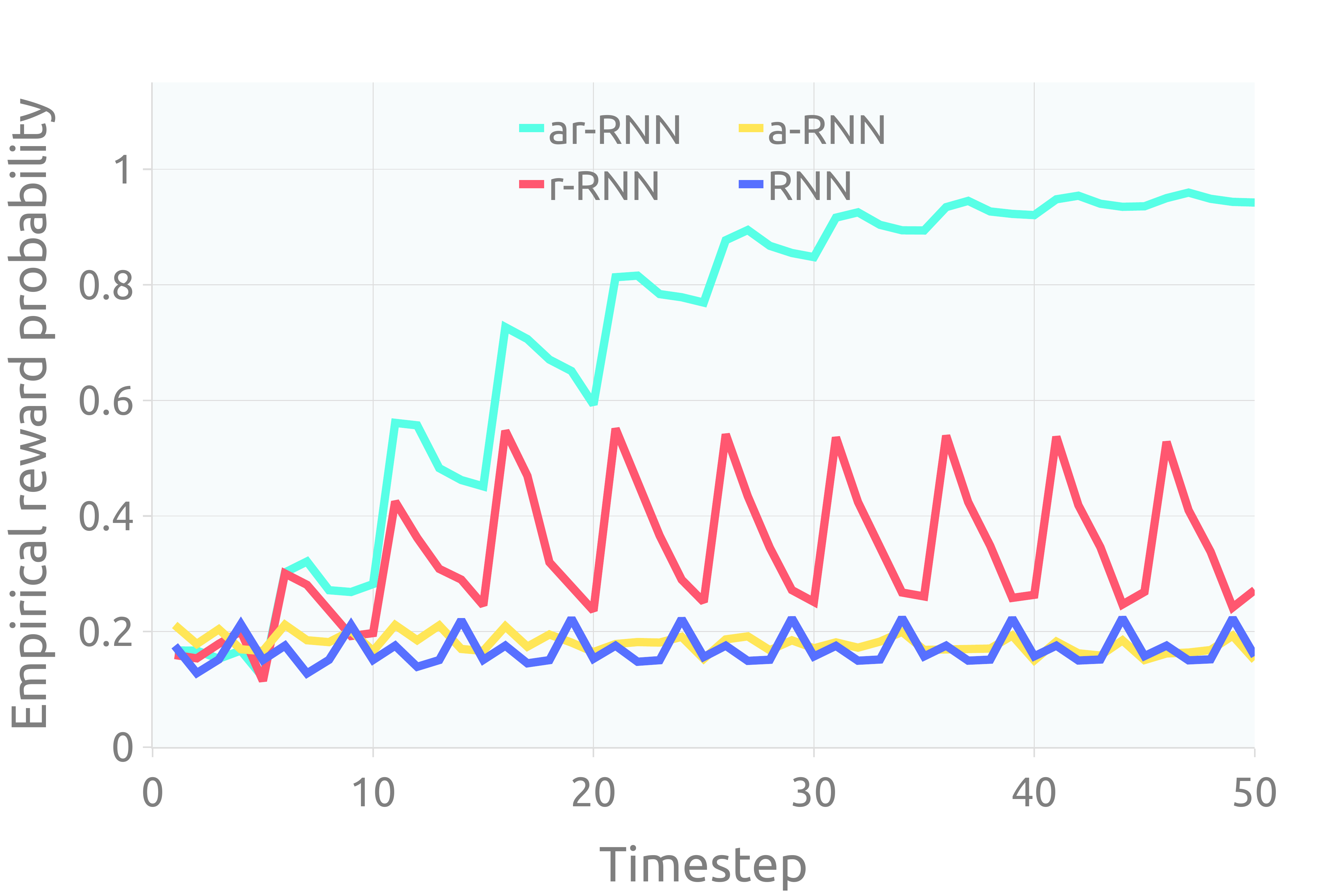}
    }
    \\
    \subfloat[DSL Game]{
        \includegraphics[trim={0cm 0cm 0cm 1cm},clip,width=0.45\textwidth]{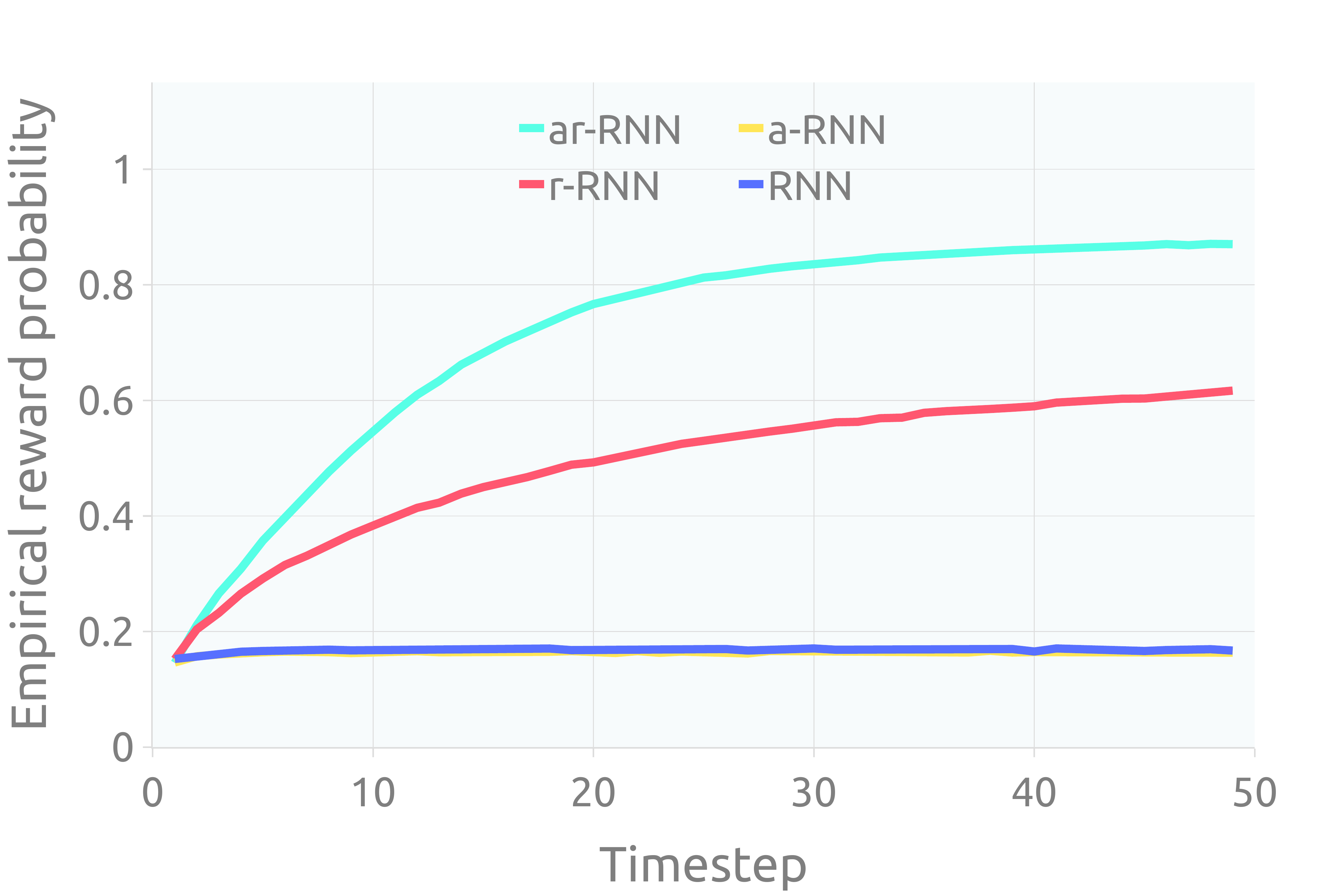}
    }
    
    \caption{\textbf{Chance of getting a reward within a trajectory.} The graph shows the chance of getting a reward at each timestep in a trajectory. The chances are calculated by averaging the reward at each timestep across all test partners and training seeds. As time goes by, ar-RNN and r-RNN learn more about their current partner and have a higher chance of getting a reward. Whereas, RNN and a-RNN do not show this behavior.}
    \label{fig:results}
\end{figure}

To answer the first question, we test meta-trained agents in the LC game and the DSL game. We evaluate four model variations, which have different input features to the GRUs (see Fig. \ref{fig:architectures}). The models are: 
\begin{itemize}
    \item RNN: Baseline recurrent neural network (RNN) with the standard observation.
    \item a-RNN: Recurrent neural network with the standard observation and the previous action signal $a_{t-1}$.
    \item r-RNN: Recurrent neural network with the standard observation and the reward signal $r_{t-1}$.  
    \item ar-RNN: Recurrent neural network with the standard observation, the previous action signal $a_{t-1}$ and the reward signal $r_{t-1}$. 
\end{itemize}

Every model variation consists of two Gated Recurrent Unit (GRU) layers \cite{cho2014learning}, each with 64 units, with two output layers: one for the policy and another for the critic. We train the model with proximal policy optimization (PPO) \cite{schulman2017proximal} with Adam optimizer \cite{kingma2014adam}. We evaluate each model using five training seeds.

As a reference, an agent with complete knowledge of the partner's pattern would theoretically get a score of 50.0 (total timesteps in an episode). The results from Fig. \ref{fig:unseen_table} show that recurrent architectures with the previous reward as an additional input (r-RNN and ar-RNN) outperform the ones that do not have the feature as input (RNN and a-RNN).
Specifically, when matched with unseen agents, ar-RNN has the highest average score of 35.03 in LC game and 34.94 in DSL game. However, RNN and a-RNN have a lower average score than a random policy that would have an expected score of 10.0.

Also, we examine the behavior of agents across timesteps within an evaluation episode. We expect that cooperative agents might struggle at the beginning when paired with an unseen agent and then perform better later on in the episode. Fig. \ref{fig:results} shows a chance of getting a reward at each timestep. The chance is calculated by averaging the reward at each timestep across all evaluation episodes from all test partners. 
We can see that both ar-RNN and r-RNN have an adaptive behavior. Especially, ar-RNN can learn to cooperate with unseen partners relatively quickly and keep improving as the episode goes on.
These experiments indicate that ar-RNN and r-RNN have cooperative skills while a-RNN and RNN do not. We interpret these results as follow: (i) The reward signal $r_{t-1}$ is necessary for the emergence of cooperative skills. This is because the agent needs to know whether or not its current strategy is suited to the current partner.
(ii) In combination with the reward signal, the previous action input $a_{t-1}$ helps the agent to cooperate better because this feature can be used by the RNN to correlate the action with the reward. This makes it easier for the RNN to identify what is the correct action during adaptation.

\begin{figure}[t!]
    \centering
    \subfloat[LC Game]{
        \includegraphics[trim={0cm 0cm 0cm 1cm},clip,width=0.45\textwidth]{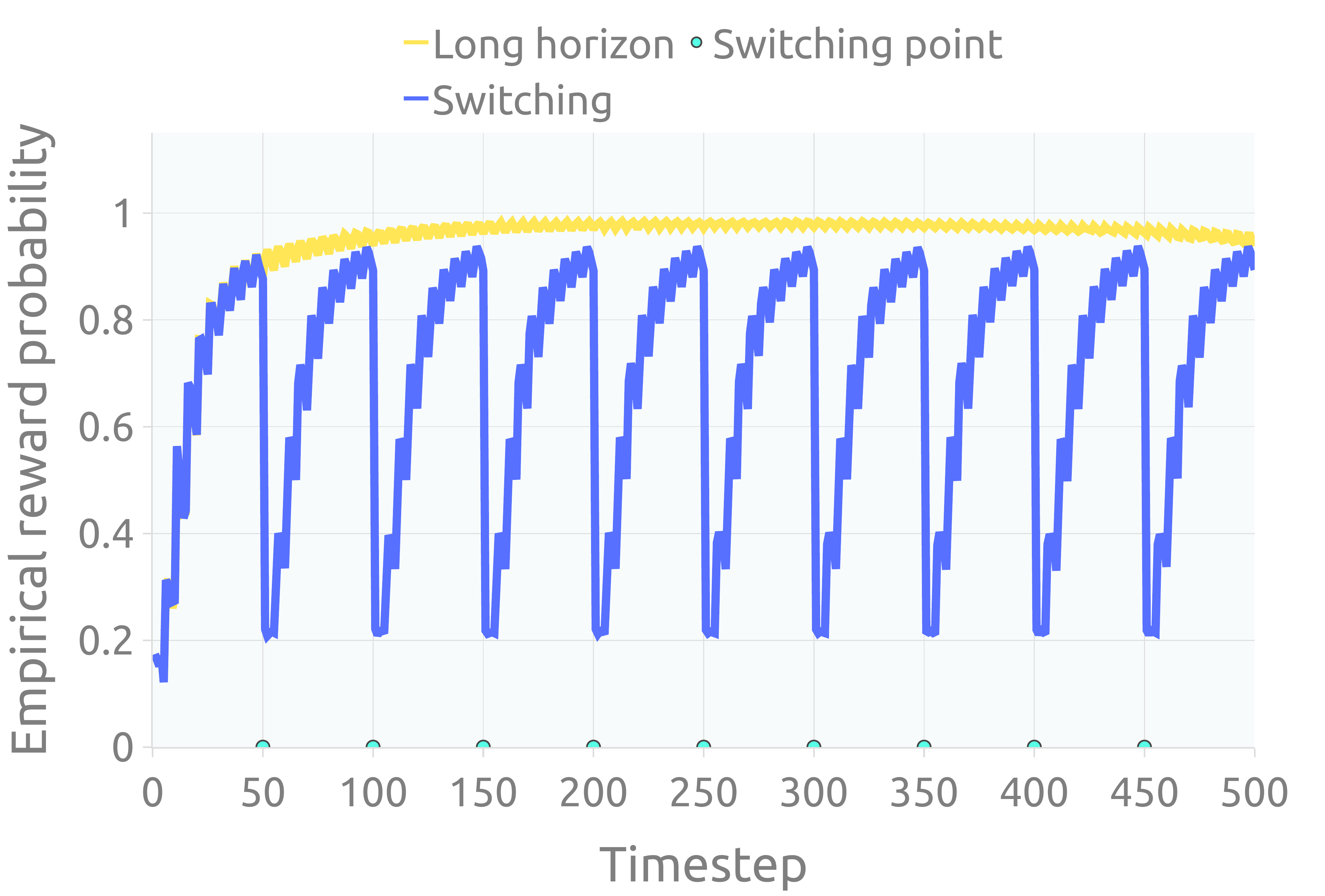}
    }
    \\
    \subfloat[DSL Game\label{overfit}]{
        \includegraphics[trim={0cm 0cm 0cm 1cm},clip,width=0.45\textwidth]{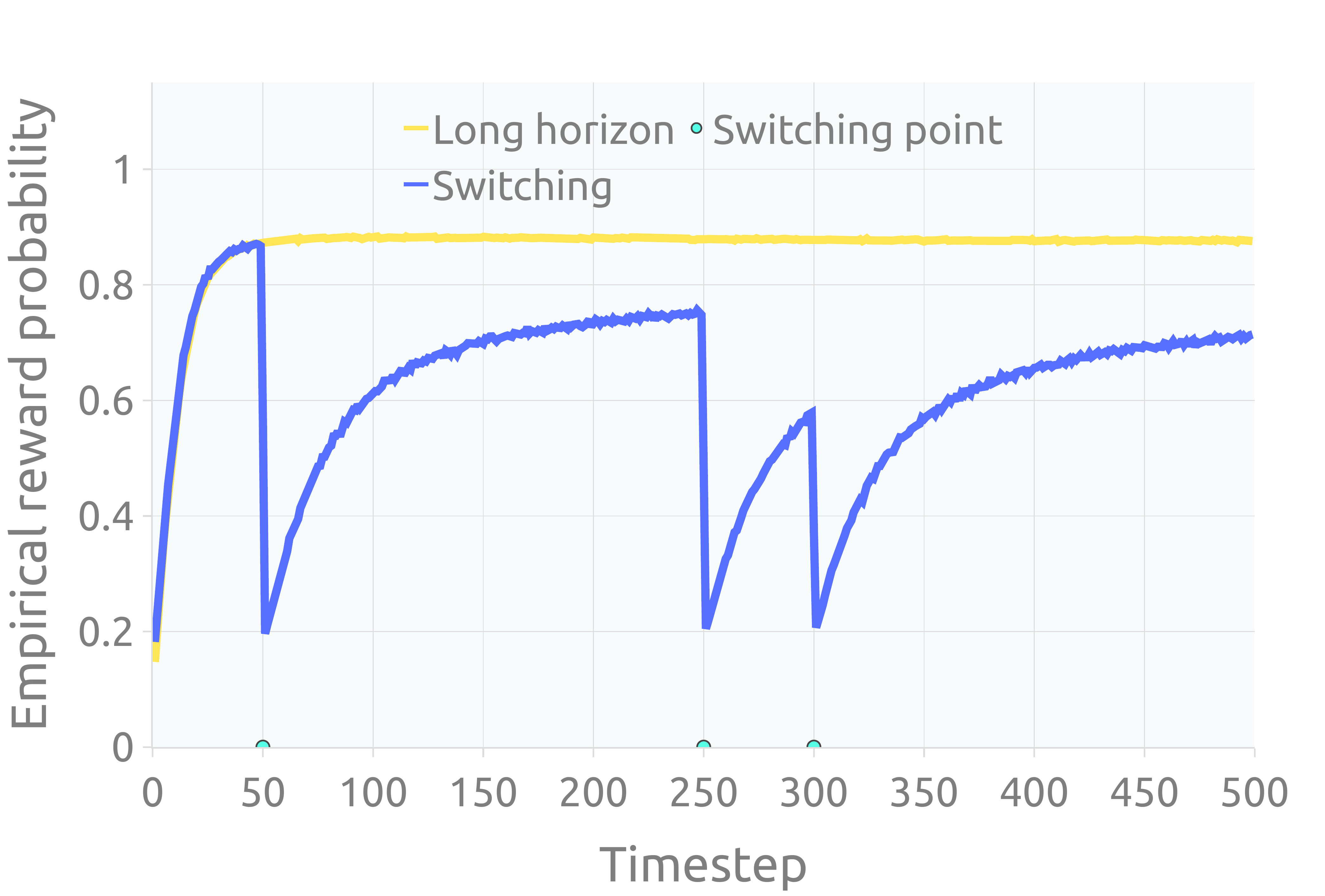}
    }

    \caption{\textbf{Continual adaptation.} We test meta-RL agents in episodes with a longer horizon and partner switching. The agents have stable performance when put into extremely long episodes. Also, they can adapt to multiple partners when the partner is changed periodically within an episode. However, we see that later adaptations are slower than the first one. We set the switching periods differently to clearly see the behavior. The periods are set to be 50 timesteps in the LC game and alternate between 50 and 200 timesteps in the DSL game. Because, in the DSL game, the agents take a much longer time to adapt than how long they do in the LC game.}
    \label{fig:continual}
\end{figure}

\subsection{Continual Adaptation}

When an agent is deployed into the real world, test-time scenarios might differ from the ones that are used during the training. In this section, we examine the performance of meta-RL agents under unexpected situations including working under a longer horizon and partner switching.

\subsubsection{Longer horizon}

When an agent is deployed into a real-world environment, identifying the end of trajectory in cooperative tasks might not be trivial. Each cooperative task might have a different horizon length given an unseen partner. For example, one agent could work slower than the others. This problem motivates our investigation into the scenario where the meta-RL agent has to cooperate with an unseen partner for longer than it expects.

In this experiment, the agent is tested in trajectories with a horizon length of 500. This is much longer compared to the horizon length of 50 that the agent is trained with. Fig. \ref{fig:continual} (Long horizon) shows that a meta-RL agent is robust when it performs to longer horizon length in both environments. The performance is stable throughout the entire trajectory.

\subsubsection{Partner Switching}
Working with only one partner throughout an entire trajectory might not be realistic when considering real-world applications where behavior or partner switching could occur over the course of the task. Here, we investigate meta-RL ability to adapt under this circumstance \emph{without explicitly trained or designed} for this situation.

In this experiment, the partner agent is changed multiple times within an evaluation episode. The switching occurs every 50 timesteps in the LC game and alternating between 50 and 200 timesteps in the DSL game to highlight the adaptation speed. The results are shown in Fig. \ref{fig:continual} (Switching). As can be seen from the results, the meta-RL agent can adapt flexibly even though it has been trained to adapt with only one partner per episode. In the DSL game, we notice a different adaptation behavior when the agent is already adapted to one partner. Specifically, it adapts much faster to the first partner compared to later partners. 

\subsection{Impact of Quantity and Diversity of Training Partners}\label{impact}
\begin{figure}[th!]
    \centering
    \subfloat[LC Game]{
        \includegraphics[trim={0cm 0cm 0cm 0.8cm},clip,width=0.45\textwidth]{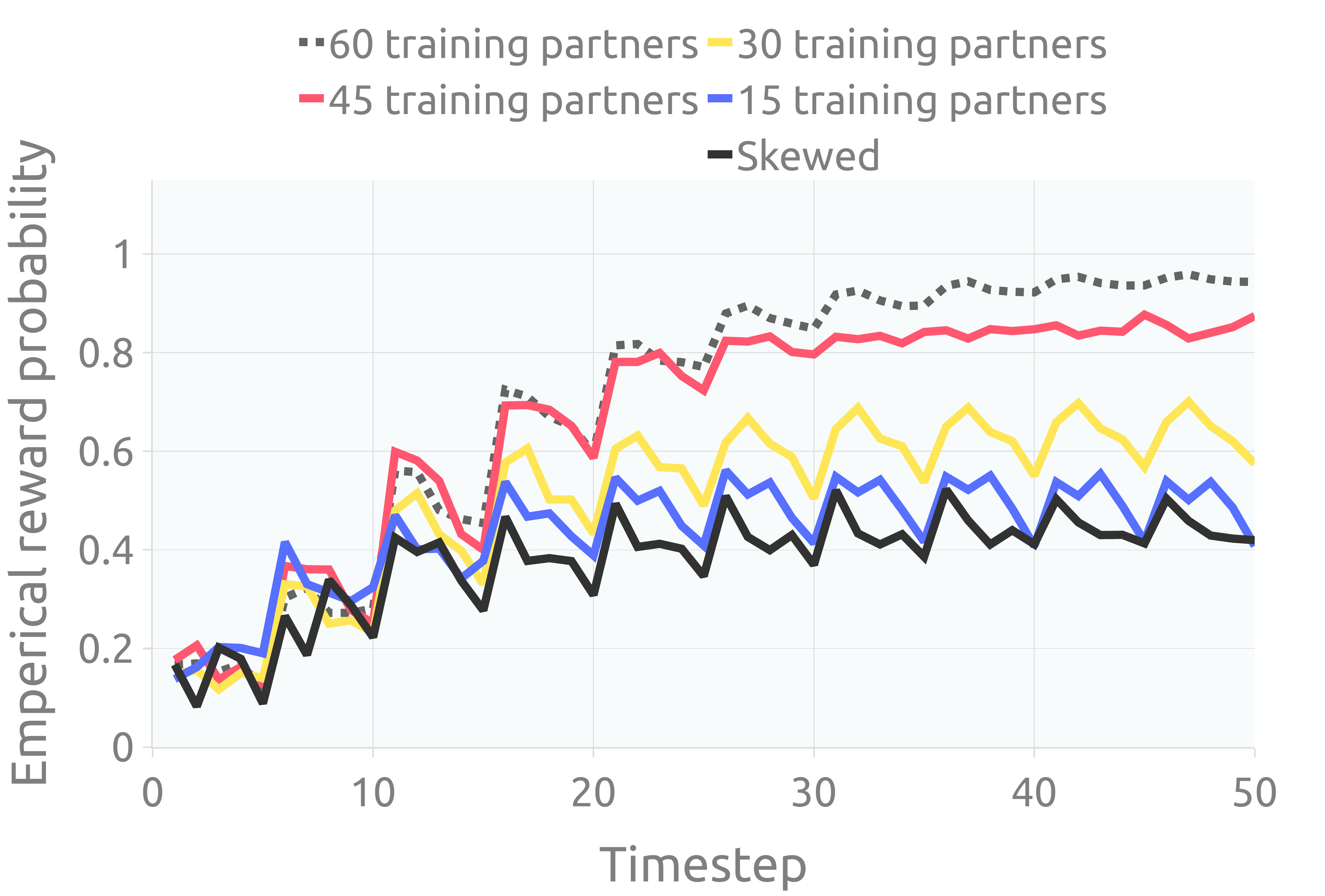}
    }
    \\
    \subfloat[DSL Game]{
        \includegraphics[trim={0cm 0cm 0cm 0.8cm},clip,width=0.45\textwidth]{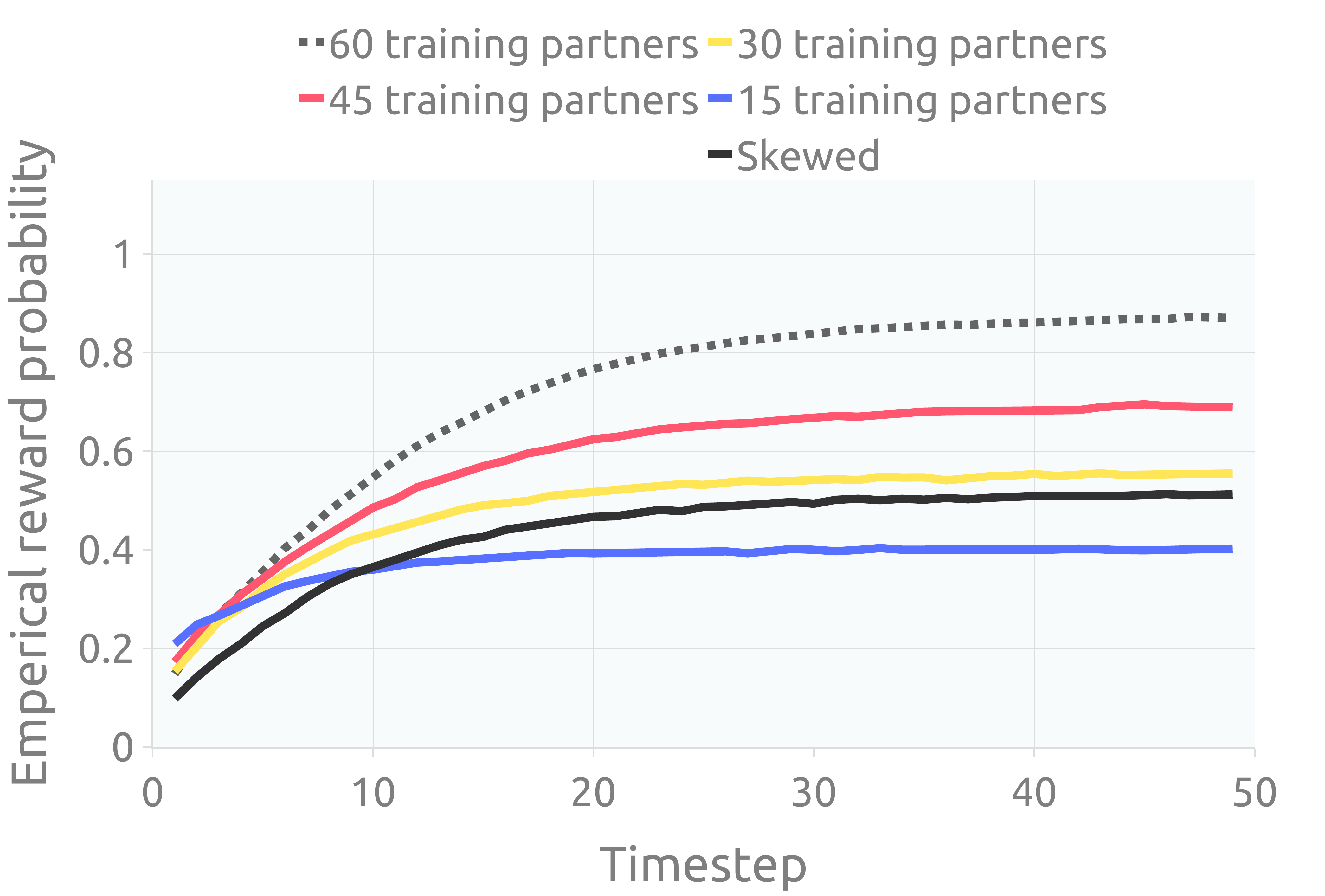}
    }

    \caption{\textbf{Performance of different training setups when matched with unseen agents.} The graph shows the impact of training distributions in terms of quantity and diversity. This emphasizes the quality and quantity of training partners when using meta-RL. The standard training distribution is shown as a reference using a dotted line.}
    \label{fig:ood}
\end{figure}

Up to this point, the experimental results have shown that meta-RL can produce robust ad hoc agents. In this experiment, we want to find the limitations of the agents and pitfalls that one needs to avoid when using this training method.

As discussed above, one of the key considerations for performing meta-training is to train the agent using a distribution over tasks or, in this work, the distribution over partner agents.
Therefore, in this experiment, we investigate the impact of the distribution over partners by varying two important properties of the training distribution: (i) quantity of partners and (ii) diversity of partners. The result for each variation is calculated over three training seeds.

\subsubsection{Quantity}
First, we study the impact of the number of training partners. In addition to the previous experiments, We consider the number of training partners from the set of \{15,30,45\}. Although the number of training partners is different, all training runs use the same amount of timesteps during training. Fig. \ref{fig:ood} shows that the generalization (i.e., cooperating with unseen partners) gets better as we increase the number of training partners. This result indicates that the quantity of training partners has a direct impact on the adaptation of the agent. Thus, when adopting meta-RL, one might need a large number of training partners available during training.

\subsubsection{Diversity} \label{diversity_test}
Next, we study the impact of diversity of the training partners. Instead of randomly selecting training partners from the pool of all possible agents, we select the training partners such that they only come from a specific part of the behavior space. In both games, a behavior can be represented with a list of numbers. For example, [3,2,4,0,1] or [2,3,1,4,0]. We use this property to define similar behaviors in each game.

\textbf{LC Game: }
a partner agent always repeats five actions. Therefore, any patterns that contain the same number sequence, e.g.  [{\color{red}1},{\color{green}2},{\color{blue}3},{\color{magenta}4},{\color{cyan}0}], [{\color{cyan}0},{\color{red}1},{\color{green}2},{\color{blue}3},{\color{magenta}4}], [{\color{magenta}4},{\color{cyan}0},{\color{red}1},{\color{green}2},{\color{blue}3}], [{\color{blue}3},{\color{magenta}4},{\color{cyan}0},{\color{red}1},{\color{green}2}], and [{\color{green}2},{\color{blue}3},{\color{magenta}4},{\color{cyan}0},{\color{red}1}] are considered to be similar. 
We also consider sequences that contain the same numbers at the same indices to be similar, e.g. [{\color{blue}2,4,3},1,0] is similar to [{\color{blue}2,4,3},0,1] but not to [3,1,0,4,2]. We then group 60 similar agents for training and use the others for evaluation.  

\textbf{DSL Game: }
the more the lists overlap the more the behaviors are similar. For instance, [{\color{blue}4,2,3},1,0] is closer to [{\color{blue}4,2,3},0,1] than [0,4,1,3,2]. This is similar to the notion of edit distance \cite{navarro2001guided}. A group of similar behaviors can be achieved by sorting because when sorted, the numbers that have the same starting numbers will be close together. Then, we use the first half of the sorted list (60 agents) as training partners and the rest as test-time partners.

This grouping skews the partner selection process such that some lever sequences or some mapping pairs will appear more often, while some are not presented during the training process at all. This is related to out-of-distribution tests in supervised learning. (The exact grouping can be found in the supplementary file.)

The training scores of meta-RL agents do not deteriorate when trained under this skewed distribution. However, Fig. \ref{fig:ood} (skewed) shows that the test-time performance reduced significantly in both environments. This result suggests that lack of diversity is detrimental to the generalization of the meta-RL agent causing the agent to be less adaptive to unseen agents during test-time.

\subsection{Working with Learned Partners}\label{adaptive_test}
\begin{figure}[t]
    \centering
    \subfloat[LC Game]{
        \includegraphics[trim={0.4cm 1cm 0.5cm 1.1cm},clip,width=0.45\textwidth]{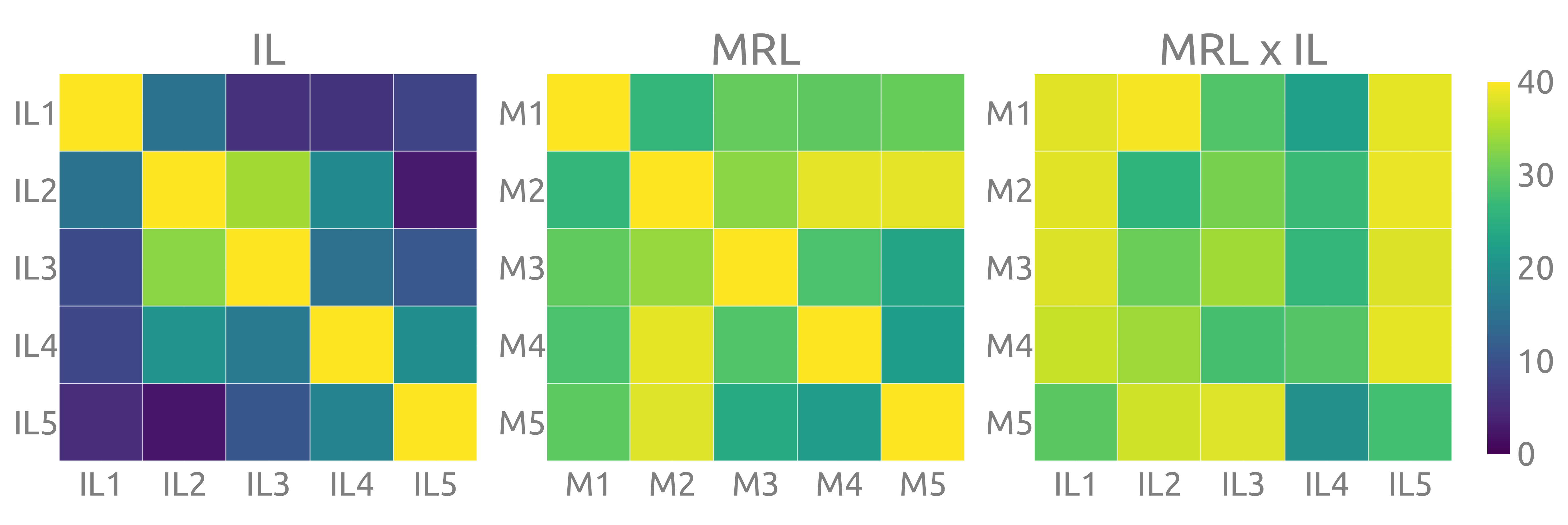}
    }
    \\
    \subfloat[DSL Game]{
        \includegraphics[trim={0.4cm 1cm 0.5cm 1.1cm},clip,width=0.45\textwidth]{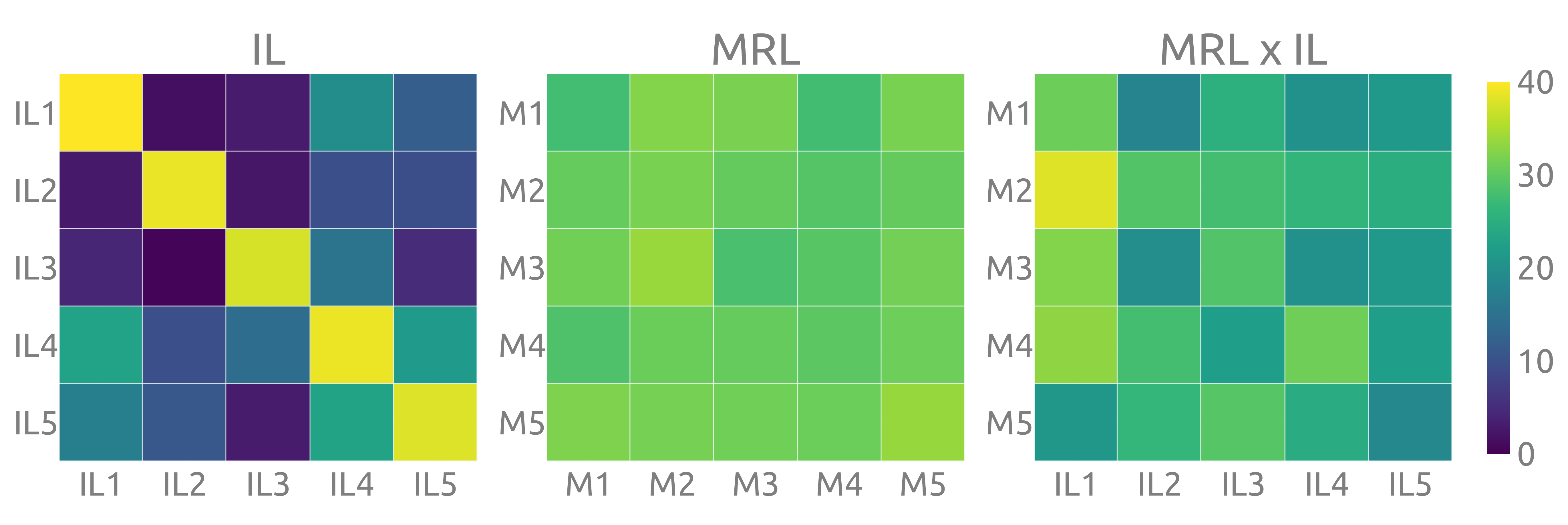}
    }

    \caption{\textbf{Cross-play matrices.} y- and x-axis represents ``player 1'' and ``player 2'' respectively. In the LC game, the role assigned is irrelevant whereas in the DSL game, ``player 1'' is a listener and ``player 2'' is a speaker. Different running numbers indicate different random seeds used in the training period. The value of each cell in the matrix represents an average evaluation score of a corresponding pair of agents using 300 episodes. Table \ref{tab:cross-play} shows quantitative results for each matrix.}
    \label{fig:cross-play}
\end{figure}

So far, we only test meta-RL agents with pre-programmed non-adaptive agents. In this experiment, we consider learned and adaptive agent as an evaluating partner. We use independent learner (IL) \cite{tampuu2017multiagent,tan1993multi} as a baseline algorithm in this experiment because it can train agents, in multi-agent environments, to achieve high performance while being simple to implement. An IL agent is trained with another concurrently learning partner. Basically, using off-the-shelf RL algorithm to optimize both agents simultaneously similar to self-play \cite{bansal2017emergent} but each agent does not share neural network weights. For clear comparisons, we use IL agents with the same architecture and hyper-parameters as meta-RL agents.


Fig. \ref{fig:cross-play} shows cross-play matrices \cite{hu2020other} for both games.
Cross-play is used as an evaluation metric for measuring the degree of generalization by evaluating multiple trained agents of the same training algorithm. In a two-player game, this can be done by pairing agents from different runs and evaluate all possible pairings.
This means the value of the diagonal cells in the IL's cross-play matrix are evaluated with the same partner that it is trained with. For meta-RL agent, they represent the performance when working with a clone of itself. The value in each off-diagonal cell represents an evaluation score when an agent is paired with an unseen partner.

As we can see, IL agents achieve high scores when evaluated with their training partner. However, they achieve low cross-play scores (i.e., performance when pairing with different runs) in both games. Meta-RL agents, on the other hand, can work with its clone and other agents from different runs thanks to their adaptability. Quantitative results are shown in Table.\ref{tab:cross-play}.

After obtaining the cross-play matrices from each method, we investigate further if meta-RL can cooperate with unseen IL agents. The rightmost matrices in Fig. \ref{fig:cross-play} show ad hoc performance when matching meta-RL agents with IL agents. In the DSL game, a meta-RL agent takes the role of a listener while an IL agent plays as a speaker. The ad hoc performance when matching meta-RL agents with IL agents is satisfactory, achieving 32.62 and 25.50 in the LC and the DSL game respectively.

These results show the robustness of meta-RL agents when working with non-stationary policies that differ from the training distribution. The robustness is beneficial when we want an agent to cooperate with non-stationary policies when designing adaptive training partners is difficult or not possible.
 
\begin{table}[t]
\caption{Cross-play and ad hoc performance of independent learner (IL) and meta-RL (MRL) agents. Seen (for IL) and clone (for MRL) score is an average of diagonal cells. The cross-play score is an average of non-diagonal cells. Ad hoc performance is calculated using an average of all cells from the rightmost matrices (MRL x IL) from Fig. \ref{fig:cross-play}}
\label{tab:cross-play}
\vskip 0.15in
\begin{center}
\begin{sc}
\scalebox{0.8}{
\begin{tabular}{c|ccc}
\toprule
\textbf{LC game}    & IL                    & MRL                       & MRL x IL                  \\ \hline
seen/clone          & $48.75 (\pm 1.01)$    & $49.28 (\pm 0.27)$        & N/A                       \\
cross-play          & $13.80 (\pm 8.59)$    & $\bm{30.02 (\pm 5.21)}$   & N/A                       \\
ad hoc              & N/A                   & N/A                       & $\bm{32.62 (\pm 5.69)}$   \\ \midrule
\textbf{DSL game}   &                       &                           &                           \\ \cline{1-1}
seen/clone          & $38.80 (\pm 1.07)$    & $30.26 (\pm 2.21)$        & N/A                       \\
cross-play          & $10.42 (\pm 7.38)$    & $\bm{30.84 (\pm 1.36)}$   & N/A                       \\
ad hoc              & N/A                   & N/A                       & $\bm{25.63 (\pm 5.09)}$   \\

\bottomrule
\end{tabular}}
\end{sc}
\end{center}
\vskip -0.1in
\end{table}

\section{Discussion} \label{discussion}
The environments used in this work have dense reward signal (i.e., an agent could get a positive reward in every timestep if the correct action is chosen). This is an easy setup for the meta-RL agent to adapt because it can correlate the previous action with the immediate feedback (reward signal). In future work, we would like to study more challenging environments with sparse rewards (e.g., Hanabi).

Also, we hypothesize that the contribution of the previous action $a_{t-1}$ in ar-RNN make it performs better than r-RNN is because of the fact that the policy implemented in this work is stochastic. The neural network could not be certain which action is taken if it outputs an action distribution that will be sampled by an external process, thus the information of the previous action taken is unknown to the network itself. The contribution of the previous action $a_{t-1}$ could be different when incorporating with deterministic policies. We leave this investigation for future work.

As shown in Fig. \ref{fig:continual}, the meta-RL agent can adapt to the first partner relatively quickly but the adaptation process becomes slower with later partners in the DSL game. We think that the RNN is optimized such that the initial state (i.e., zero vector) is used as an indicator that it is interacting with a new partner. But, under the partner switching scenario, the RNN's internal states are not reset when the partner is changed. Hence, the RNN adapt slower without the indicator. Interestingly, this behavior is only displayed in the DSL game and not the LC game.

Partner switching has been studied by \citet{ravula2019ad}. They formulate the problem of behavior switching agents as the Change Point Detection problem. We expect that a Change Point Detection algorithm could be beneficial when used in combination with meta-RL. The algorithm could predict when the RNN should reset its internal state to forget about the previous partner's behavior and readily adapt to a new partner.


Section \ref{impact} displays potential issues that could occur if the training distribution is not properly set up for meta-RL agents. In multi-agent environments, acquiring a substantial amount of training partners with diversity might not be trivial in some environments. Especially in high-dimensional environments that an agent's behavior cannot be hand-coded or procedurally generated. This problem has been hindering the progress of research in the multi-agent domain as we see only a fraction of work that considers high-dimensional environments. Most work, including ours, rely heavily on the fact that they could generate such training partners \cite{canaan2020generating,ghosh2019towards,he2016opponent,shih2021critical,zheng2018deep_bpr+}. All in all, we think that obtaining a substantial diverse set of training partners is still a challenging task and is a worthwhile research question to be investigated \cite{lupu2021trajectory, strouse2021collaborating}.

From these results, we believe that meta-RL is a feasible approach for training a cooperative agent that can work with unseen agents. Although, in this work, we only consider meta-RL in its original form proposed by \citet{duan2016rl,wang2016meta_rl}. It is conceivable that techniques from the multi-agent domain (e.g., centralized critic) could be applied to meta-RL when training a cooperative agent with meta-RL in more complex environments.


\section{Conclusion} \label{conclusion}
This paper demonstrates that meta-RL recipes could be used to produce cooperative agents that work well under the ad hoc teamwork problem. Additionally, we identify that the reward signal is the key component for the emergence of adaptive and cooperative skills. 

Meta-RL agents also show robustness under situations they have never seen before during training, including extremely longer horizon and partner switching. However, the quantity and diversity of training partners have a direct impact on the generalization of meta-RL agents. Also, we show that if the training is not setup properly, meta-RL agents could also be very brittle and could not generalize to unseen agents. Finally, we show that meta-trained agents can coordinate with each other and work with unseen IL agents despite only trained with fixed behavior partner agents.



\bibliography{ref}
\bibliographystyle{icml2021}
\end{document}